\documentclass{article} 

\usepackage{Apriel_Nemotron_15B}
\usepackage{comment}
\usepackage{tabularx}
\usepackage{multicol}
\usepackage{amsmath}
\usepackage{amssymb}
\usepackage{amsfonts}

\usepackage[utf8]{inputenc} 
\usepackage[T1]{fontenc}    
\usepackage{newunicodechar}
\newunicodechar{ }{\,} 
\usepackage{hyperref}       
\usepackage{url}            
\usepackage{booktabs}       
\usepackage{amsfonts}       
\usepackage{nicefrac}       
\usepackage{microtype}      
\usepackage{xcolor}         
\usepackage{makecell}
\usepackage{tikz}
\usepackage{pgfplots}
\usepackage{graphicx}
\usepackage{subcaption}
\pgfplotsset{compat=1.18}
\pgfplotsset{scaled ticks=false}

\usepackage[
  backend=biber,
  style=numeric,
  sorting=none
]{biblatex}

\bibliography{Apriel_Nemotron_15B}

\title{Apriel--H1: Towards Efficient Enterprise Reasoning Models}

\author{\textbf{SLAM Lab}\\ ServiceNow\thanks{Correspondence to  \href{mailto:oleksiy.ostapenko@servicenow.com}{oleksiy.ostapenko@servicenow.com}, contributors are listed in Section~\ref{sec:contributors}}}

\begin{document}

\maketitle
\begin{figure}[h]
    \centering
    \includegraphics[width=0.4\textwidth]{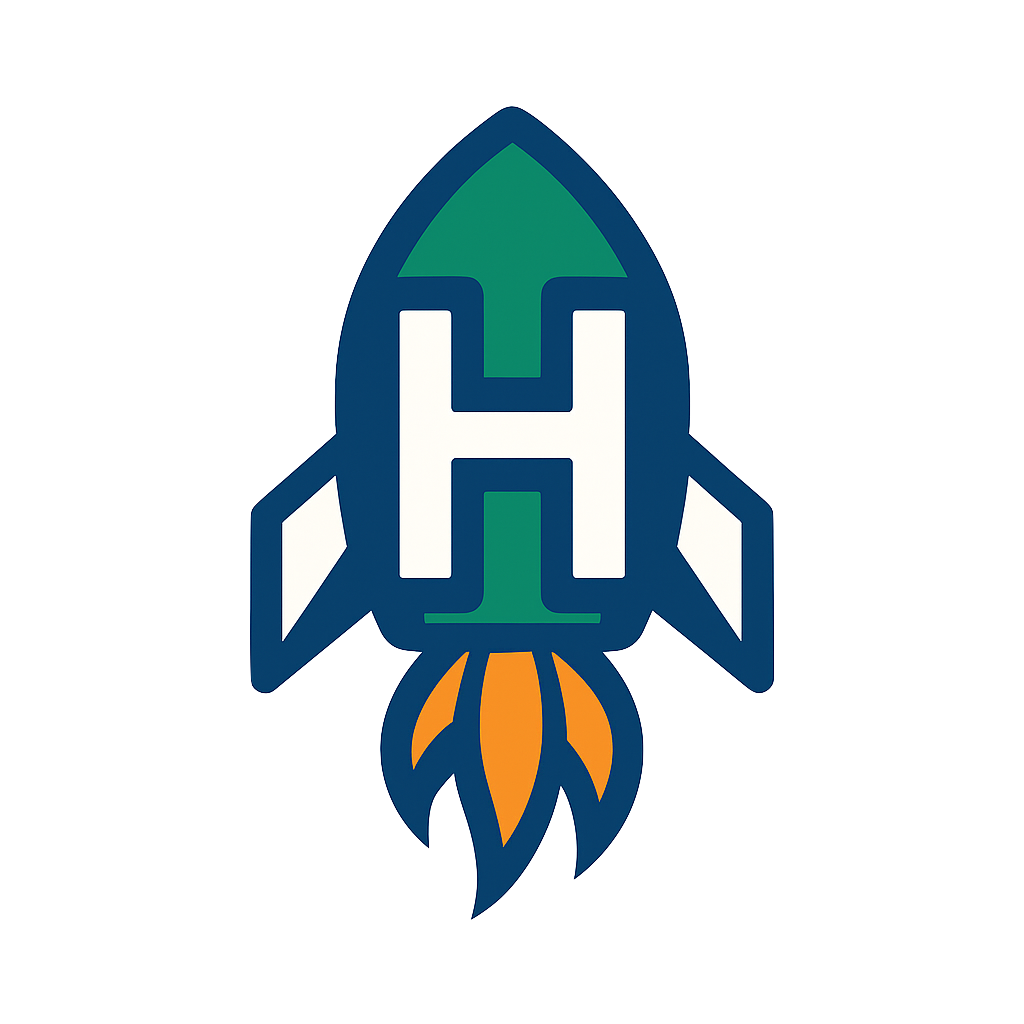}
\end{figure}

\begin{abstract}
    Large Language Models (LLMs) achieve remarkable reasoning capabilities through transformer architectures with attention mechanisms. However, transformers suffer from quadratic time and memory complexity in the attention module (MHA) and require caching key-value states during inference, which severely limits throughput and scalability. High inference throughput is critical for agentic tasks, long-context reasoning, efficient deployment under high request loads, and more efficient test-time compute scaling.
    
    State Space Models (SSMs) such as Mamba offer a promising alternative with linear inference complexity and a constant memory footprint via recurrent computation with fixed-size hidden states. In this technical report  we introduce the \emph{Apriel-H1} family of hybrid LLMs that combine transformer attention and SSM sequence mixers for efficient reasoning at 15B model size. These models are obtained through incremental distillation from a pretrained reasoning transformer, Apriel-Nemotron-15B-Thinker, progressively replacing less critical attention layers with linear Mamba blocks.
    
    We release multiple post-distillation variants of \emph{Apriel-H1-15B-Thinker} with different SSM-to-MHA ratios and analyze how reasoning performance degrades as more Mamba layers replace MHA. Additionally, we release a 30/50 hybrid variant of Apriel-H1, further fine-tuned on a supervised dataset of reasoning traces, achieving over 2× higher inference throughput when deployed in the production-ready vLLM environment, with minimal degradation in reasoning performance. This shows that distilled hybrid SSM–Transformer architectures can deliver substantial efficiency gains over the pretrained transformer equivalent without substantially compromising the reasoning quality.

\end{abstract}

\newpage
\section{Introduction}

The transformer architecture \cite{vaswani2017attention} has become the de facto standard for large-scale language modeling (LLMs), powering state-of-the-art models such as recently released Apriel models \cite{Apriel-nemotron-15b-thinker,Apriel-small-language-models,radhakrishna2025apriel1515bthinker}. The throughput of transformer inference models is largely limited due to the quadratic complexity of the attention module, as well as the necessity to cache and retrieve key and value representations of preceding tokens into fast GPU memory during each forward computation~\cite{vaswani2017attention}. 

These throughput constraints can become a critical bottleneck in the practical adoption of LLMs, particularly in scenarios with high request loads, multi-user environments, as well as for tasks that require the model to consume large prompts and generate long output traces. The latter is especially relevant for argentic tasks with long contexts and reasoning traces. Maintaining high inference throughput is crucial for several reasons: (i) transformer inference is memory bandwidth-bound, leading to underutilized compute resources; (ii) high throughput enables cost-effective serving of more concurrent users; (iii) reduced latency dramatically improves user experience in business-critical interactive applications; (iv) last but not least, higher throughput can be transferred into improved performance through more extensive RL fine-tuning or more efficient test-time compute scaling~\cite{wang2025m1}.



State Space Models (SSM) such as Mamba~\cite{mamba,dao2024transformers} have been proposed as an efficient attention-free alternative to transformers, allowing for much higher inference throughput and lower per-user latency. While several recent works have shown competitive or superior performance of SSM-Transformer hybrids in language modelling and knowledge recall, obtaining competitive hybrids that can excel at complex reasoning remains challenging. In this work, we focus on distilling a pre-trained full transformer reasoning model Apriel-Nemotron-15b-Thinker~\cite{Apriel-nemotron-15b-thinker} into a transformer-SSM hybrid reasoner -- the \emph{Apriel-H1-15B-Thinker} family of models. We release multiple Apriel-H1 models with different fractions of SSM to attention layers. Our best performing model more than doubles the throughput as compared to the full transformer baseline with minimal loss in reasoning ability. While other hybrids that excel at reasoning exist (M1 \cite{wang2025m1}, Nemotron-Nano-v2 \cite{basant2025nvidia}), they are either not distilled directly from a reasoning teacher~\cite{wang2025m1} or trained completely from scratch as hybrids~\cite{basant2025nvidia}. To the best of our knowledge, the \emph{Apriel-H1-15B-Thinker} is the first open-weights SSM-Transformer hybrids that is directly \textbf{distilled} from a reasoning teacher.




\section{Background and Related works}
\label{sec:related_work}
Recurrent token mixers such as Mamba~\cite{gu2023mamba} and DeltaNet~\cite{yang2024parallelizing} have emerged as efficient alternatives to transformers, achieving linear inference complexity while maintaining competitive performance. One of the most prominent representatives of linear sequence mixers is Mamba~\cite{gu2023mamba}, which introduced selective SSMs that modulate state transitions based on input. The core Mamba computations are:

\begin{align}
h_t &= \mathbf{A_t} h_{t-1} + \mathbf{B}_t x_t \\
y_t &= \mathbf{C}_t h_t
\end{align}
    
where $\mathbf{A}_t$, $\mathbf{B}_t$, and $\mathbf{C}_t$ are dynamically computed matrices conditioned on the input at each time step. These mechanisms yield performance competitive with transformers while maintaining constant memory complexity during autoregressive decoding.

Recent hybrid approaches combine Multi-Head Attention (MHA) and recurrent sequence mixers to leverage the complementary strengths of both paradigms~\cite{waleffe2024empirical,dong2024hymba}. When trained from scratch, such hybrid architectures have been shown to outperform homogeneous models (pure attention or pure SSM) on general understanding and knowledge-recall benchmarks, while significantly improving inference throughput~\cite{lieber2024jamba,bae2025hybrid,wang2025systematic,blakeman2025nemotron}. However, achieving transformer-level performance on complex reasoning benchmarks (e.g. AIME and MATH-500) remains challenging.

A common strategy for model hybridization is knowledge distillation (KD), where a pre-trained transformer is first converted into a more efficient hybrid model and then trained using KD objective \cite{bick2025llamba,wang2024mamba,bick2024transformers}. In this setting, substantial post-distillation fine-tuning is often required to close the gap to the teacher model’s downstream performance~\cite{wang2025m1,wang2024mamba,gu2025jet}.
Here, we release several post-distillation reasoning models prior to any additional fine-tuning. These models remain amenable to further task-specific adaptation. This is exemplified by our Apriel-H1-30/50-15B-Thinker-SFT (H1-30-SFT) model, which demonstrates strong reasoning performance after supervised fine-tuning on additional reasoning traces.
Llamba~\cite{bick2025llamba} employs MOHAWK~\cite{bick2024transformers} in a three-staged procedure where layers are first distilled independently and subsequently fine-tuned through end-to-end distillation. Zebra~\cite{yang2025zebra} integrates SSMs with MLA-attention~\cite{liu2024deepseek}, guided by layer-importance heuristics based on the L2 distance between teacher and student representations.
In contrast, our approach determines layer placement using downstream performance degradation or directly using the distillation loss (see Section~\ref{sec:layer_placement}).

A simpler and increasingly popular hybridization route, {Mamba-in-LLaMA (MIL)~\cite{wang2024mamba}, 
initializes newly introduced Mamba layers directly from pretrained MHA weights using a 
\emph{linearized} view of attention (with the softmax removed). 
The resulting initialization maps the attention projections $(W_Q, W_K, W_V, W_O)$ 
to the Mamba state-space parameters as

\begin{align}
\mathbf{B}_t = W_K \mathbf{o}_t, \quad 
\mathbf{C}_t = W_Q \mathbf{o}_t, \quad 
\mathbf{x}_t = W_V \mathbf{o}_t, \quad
\mathbf{y}_t &= W_O^{\!\top}\mathbf{y}_t ,
\end{align},

where $\mathbf{o}_t$ denotes the input activations of the replaced Transformer block. 
In practice, the pretrained attention weights $W_Q, W_K, W_V$ are 
copied into the corresponding input-projection slices of the Mamba mixer 
($\mathbf{C}\!\leftarrow\!W_Q$, $\mathbf{B}\!\leftarrow\!W_K$, $\mathbf{x}\!\leftarrow\!W_V$), 
while $W_O$ initializes the Mamba output projection. To comply with the GQA nature of the teacher's MHA block, $\mathbf{B}$ and $\mathbf{x}$ are expanded by repeating them as many times as there are groups in the teachers GQA to match the required dimensionality.


In preliminary experiments, we observed no significant advantage from the MOHAWK multi-stage procedure and therefore adopted MIL initialization to avoid the complexity of multi-stage, layer-wise distillation pipeline.

Beyond Mamba, other efficient architectures such as RetNet~\cite{sun2023retentive}, RWKV~\cite{peng2023rwkv}, and DeltaNet~\cite{yang2024parallelizing,yang2024gated} similarly aim to overcome the scalability limitations inherent in attention-based models. In this release, we intentionally employ a variant of Mamba-1~\cite{mamba} as our linear mixer for its strong empirical performance across non-reasoning hybrids~\cite{lieber2024jamba,team2024jamba} and distilled architectures~\cite{wang2024mamba}. Future iterations of the Apriel-H models will explore more advanced linear mixers.
To handle the grouped-query attention (GQA) used in the original MHA layers, we follow the Mamba implementation introduced in M1~\cite{wang2025m1}, $\mathbf{B}_t$ and $\mathbf{x}_t$ are \emph{expanded} to match the total number of attention heads dictated by $\mathbf{C_t}$.

Our distillation process uses the reverse-KL formulation, similar to that used in M1~\cite{wang2025m1}, which also explored reasoning-oriented transformer-to-Mamba distillation but at a smaller scale. Preliminary experiments confirmed that reverse-KL distillation outperformed the standard forward-KL objective. Future work will investigate integrating DeltaNet and other linear attention variants into our distillation framework to further enhance the reasoning efficiency of large-scale reasoning models.

\section{Transformer to hybrid conversion}
Apriel-H1 models were distilled from Apriel-Nemotron-15B-Thinker~\cite{Apriel-nemotron-15b-thinker} models following an iterative procedure over the following stages until the desired number of MHA mixers are replaced: (1) layer importance estimation followed by (2) replacement of the MHA mixers in the $m$ least important transformer blocks with linear Mamba mixer, and (3) KD stage that aligns the hybrid model $\mathcal{M}_H$ with the original transformer teacher $\mathcal{M}_T$. In the following we describe each step in detail. We denote the resulting hybrid model using the shorthand \texttt{H1}-$h/L$, or simply \texttt{H1}-$h$, indicating that $h$ out of $L$ transformer layers are replaced with Mamba sequence mixers with $L=50$ for the teacher model Apriel-Nemotron-15B-Thinker.

\subsection{Layer importance estimation \& selection}
\label{sec:layer_placement}
We consider two procedures for estimating layer importance. The resulting importance scores are used to guide the mixer replacement (MHA → Mamba) in the pretrained model, where layers are replaced in the order of increasing importance.

\paragraph{Leave-one-out (LOO)}
For each layer $l \in {1, \dots, L}$, we construct a modified teacher model $\mathcal{M}_T^{(-l)}$ where layer $l$ is replaced with an identity function, and evaluate it on a held-out subset of downstream tasks (without fine-tuning). The importance score $I_l$ is defined as:
\begin{equation}
I_l = \mathcal{P}(\mathcal{M}_T) - \mathcal{P}(\mathcal{M}_T^{(-l)})
\end{equation}






where $\mathcal{P}(\cdot)$ denotes the performance metric, in our case downstream accuracy on MMLU \cite{hendrycks2009measuring}. A higher score $I_l$ indicates that layer $l$ contributes more to overall performance. We rank all layers by importance and select the $k$ least important ones for SSM replacement. We denote this layer placement method as \texttt{LOO} (leave-one-out). We plot layer importance for Apriel-Nemotron-15B-Thinker model in Fig.\ref{fig:layer_importance}, which is similar in trend to the layer importance reported in \cite{yang2025zebra}. 
We used LOO for the initial layer importance estimation (for H1-25/50), where we needed to estimate importance of all 50 layers, because it is cheaper (does not require training). In subsequent hybridization iterations we applied MMR estimation, which is described next.

\begin{figure}[h]
    \centering
    \includegraphics[width=0.5\linewidth]{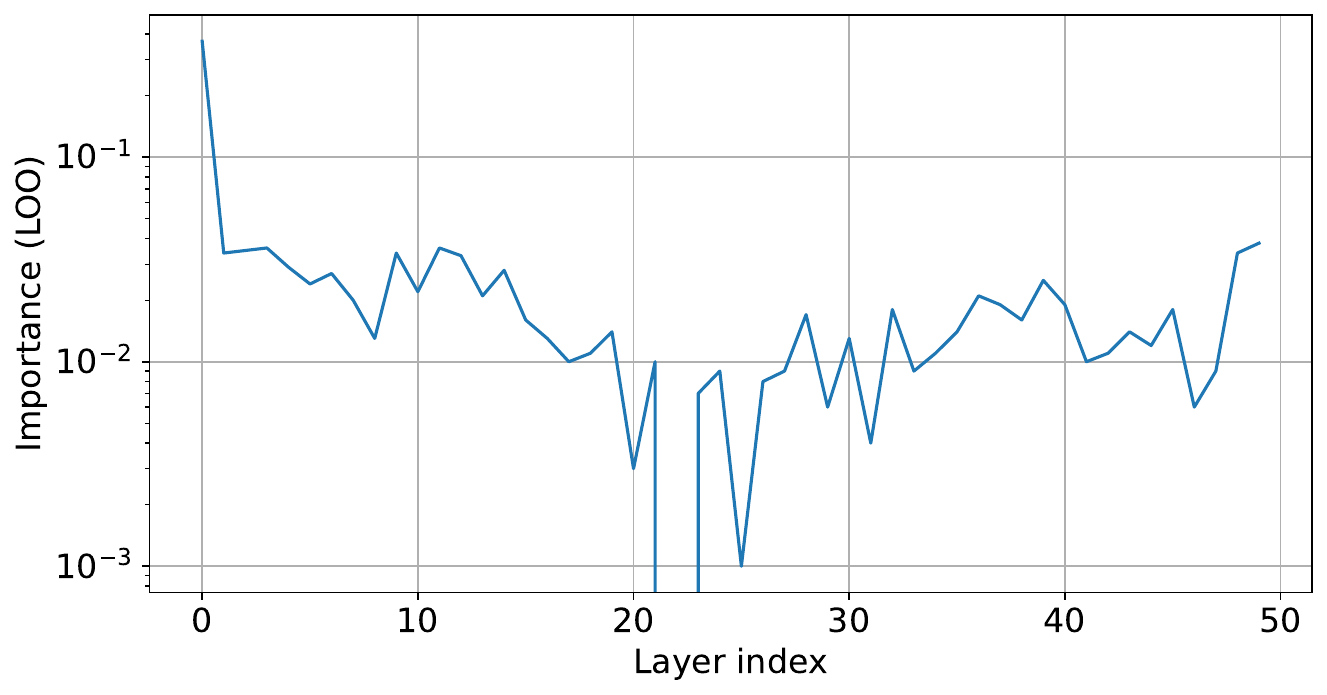}
    \caption{Layer importance ($\uparrow$) using \texttt{LOO} for the Apriel-Nemotron-15B-Thinker model~\cite{Apriel-nemotron-15b-thinker}.}
    \label{fig:layer_importance}
\end{figure}

\begin{figure}[htbp]
    \centering
    \begin{subfigure}{0.32\textwidth}
        \centering
        \includegraphics[width=\textwidth]{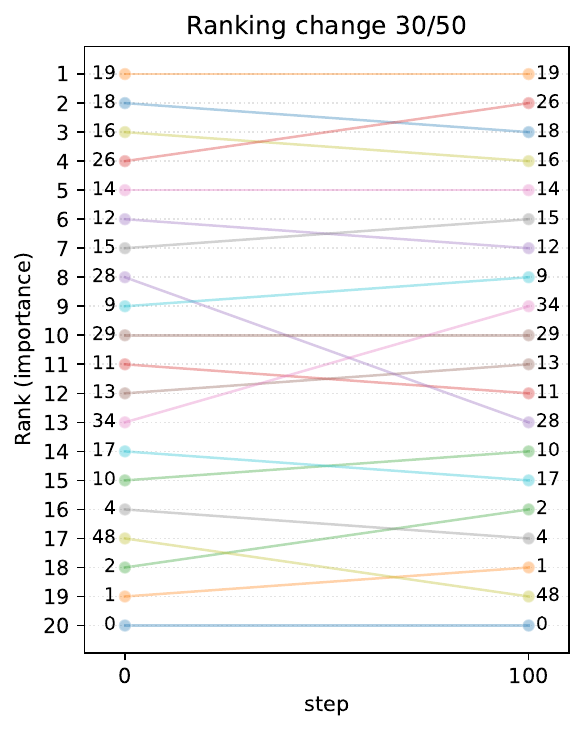}
        \caption{H1-30/50}
        \label{fig:30}
    \end{subfigure}
    \begin{subfigure}{0.32\textwidth}
        \centering
        \includegraphics[width=\textwidth]{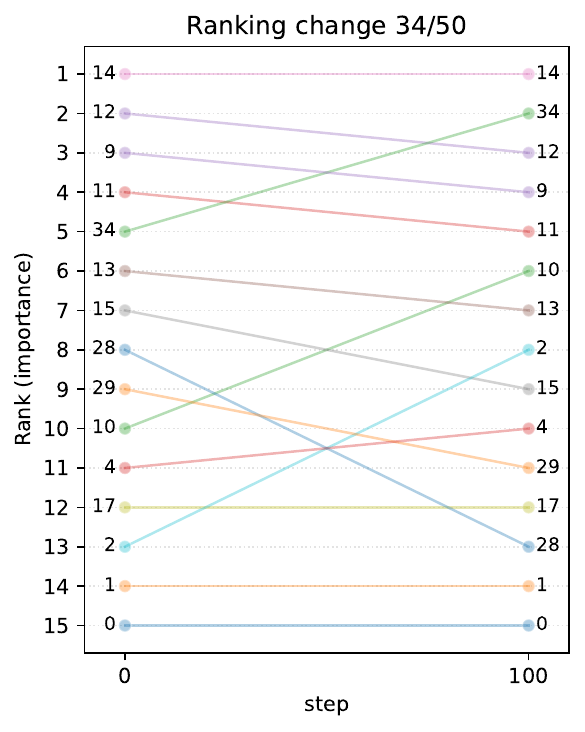}
        \caption{H1-34/50}
        \label{fig:34}
    \end{subfigure}
    \begin{subfigure}{0.32\textwidth}
        \centering
        \includegraphics[width=\textwidth]{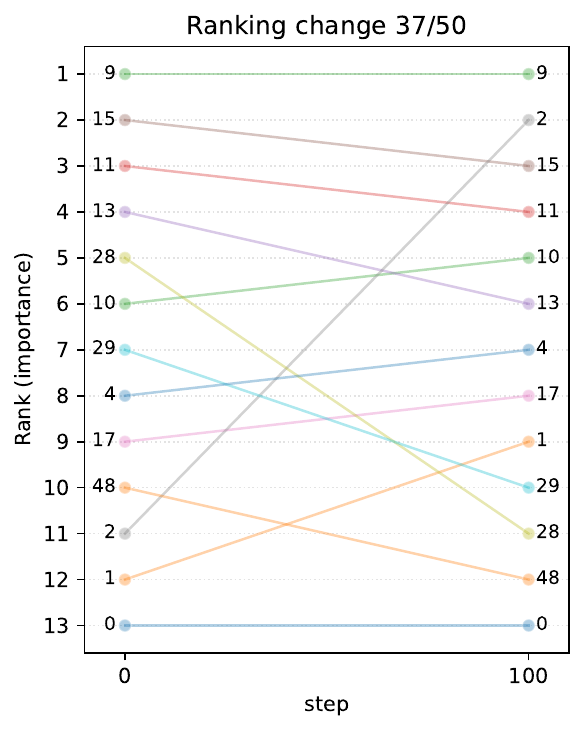}
        \caption{H1-37/50}
        \label{fig:37}
    \end{subfigure}
    \caption{Layer importance \texttt{MMR} ($\uparrow$) before distillation (0 steps) and after 100 distillation steps. Crossing horizontal lines visualize the change in layer importance ranking.}
    \label{fig:layer_importance_mmr}
\end{figure}

\paragraph{MIL-Mamba-Replacement (MMR)}
For each transformer layer $l$ in the initial model $\mathcal{M}$, which may be either a full transformer or a hybrid, we construct a modified model $\mathcal{M}^{(l \rightarrow M)}$ where the MHA mixer of layer $l$ is replaced with a Mamba sequence mixer initialized using \texttt{MIL}. We then perform a short distillation run of 100 steps (see Section~\ref{sec:distillation}) and record the final distillation loss $\mathcal{L}_{\text{KD}}$ of the trained model. This final loss is used as the \emph{layer importance score}, defined as $I_l^{\text{MMR}} = \mathcal{L}_{\text{KD}}(\mathcal{M}^{(l \rightarrow M)})$.
Intuitively, layers for which replacing the MHA mixer with a Mamba one yields a lower final distillation loss are deemed less ``important'' and are replaced earlier. We rely on the final distillation loss (after 100 steps) rather than the initial loss to capture the effect of training dynamics, since, as shown in Fig.~\ref{fig:layer_importance_mmr}, a layer may start with a higher initial loss but converge to a lower value after optimization. 

To obtain the initial hybrid model $\mathcal{M}_H$ for the next end-to-end distillation stage, we select top $k$ \emph{least} important transformer layers according to the importance score $I$ and replace their mixers with MIL initialized Mamba mixers.

\subsection{Staged end-to-end distillation}
\label{sec:distillation}
Once the target layers are replaced, we align the hybrid model $\mathcal{M}_H$ with the original transformer $\mathcal{M}_T$ via KD. We train $\mathcal{M}_H$ to minimize the reverse KL divergence between the teacher and student output distributions:

\begin{equation}
\mathcal{L}_{\text{KD}} = \mathcal{D}_{\text{KL}}\left( \text{softmax}(z_H / \tau)| \text{softmax}(z_T / \tau) \right)
\end{equation}

where $z_T$ and $z_H$ are the logits from the teacher and hybrid models, and $\tau=1$ is the softmax temperature. We default to using reverse-KL as it resulted in improved performance in our preliminary experiments.

We found that staged distillation, in which subsets of layers are distilled incrementally based on their estimated importance, outperformed single-shot replacement of all target layers. Specifically, to obtain the \texttt{H}-40/50 hybrid model, we began by distilling the \texttt{H}-25/50 hybrid variant, with layer placement determined using the \texttt{LOO} heuristic. Before each subsequent stage, we re-estimated layer importance using the \texttt{MMR} heuristic and progressively distilled additional layers following the schedule

\[
\texttt{H1-}25/50 \rightarrow \texttt{H1-}27/50 \rightarrow \texttt{H1-}30/50 \rightarrow \texttt{H1-}34/50 \rightarrow \texttt{H1-}37/50 \rightarrow \texttt{H1-}40/50.
\]

This iterative approach allowed each newly introduced SSM mixer to adapt while preserving previously distilled knowledge, resulting in smoother convergence and improved performance compared to one-shot distillation.



\begin{figure}[h!]
    \centering
    \includegraphics[width=\linewidth]{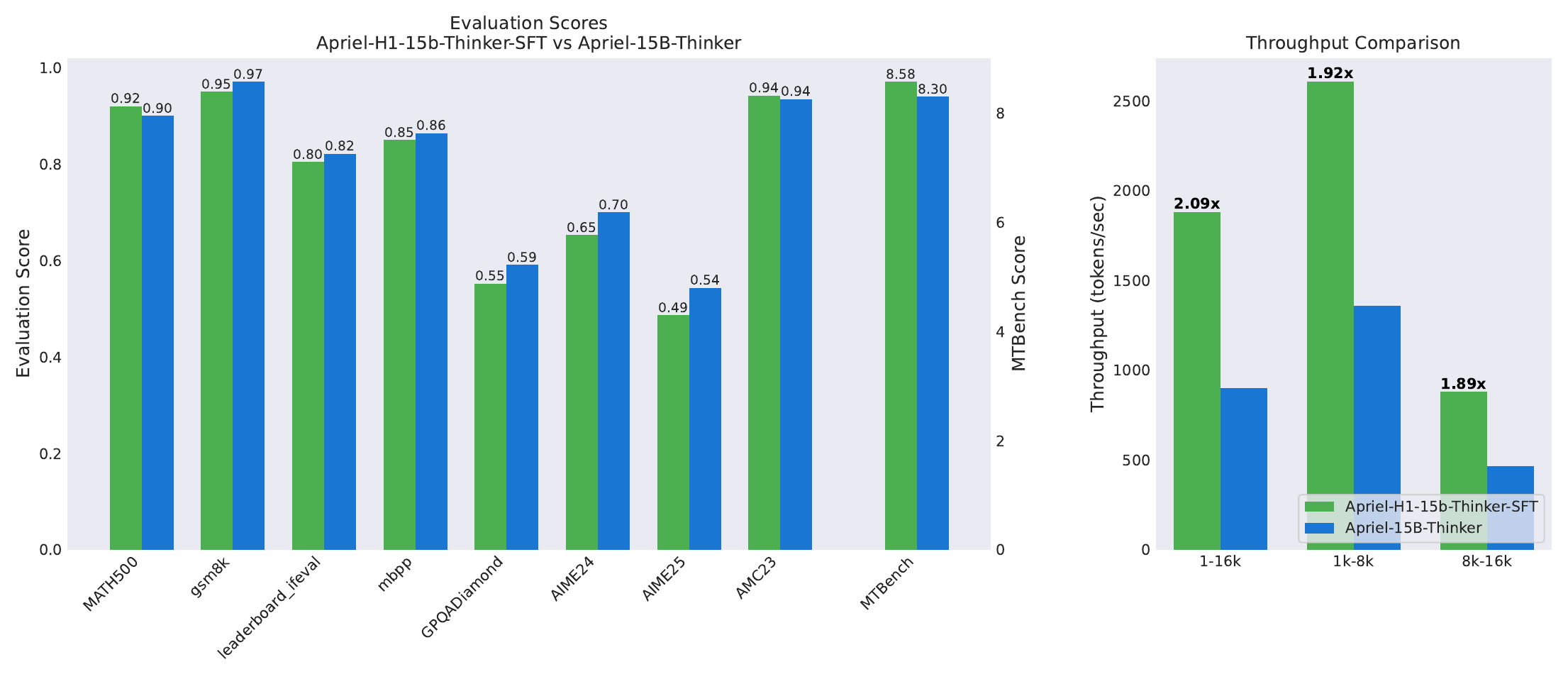}
    \caption{(left) Comparison of evaluation metrics between Apriel-Nemotron-15b-Thinker vs. Apriel-H1-30/50-15b-Thinker-SFT. (right) The \texttt{H} variant more than doubles the throughput without with minimal drop in performance across a wide range of tasks under a typical reasoning load using vLLM back-end.}
    \label{fig:eval_hybrid_transformer}
\end{figure}

\subsection{Training details}
Following the recommendation from ~\cite{wang2025m1} we use supervised fine tuning (SFT) data for model distillation. More specifically we use a subset of SFT data that was used to train the teacher model Apriel-Nemotron-15B-Thinker~\cite{Apriel-nemotron-15b-thinker} which includes $\sim$9B tokens of high-quality reasoning traces including tasks such as advanced mathematics, coding, science etc.. Each distillation phase is run with sequence length of 16,384 with batch-size of 64 and base learning rate of 5e-5 with linear decay and warm-up. The number of tokens used for each training phase is revealed in Fig.~\ref{fig:performance}. We use Fast-LLM~\cite{Lamy_Poirier_Fast_LLM_2024} training library for all training runs. For the Mamba mixers we use the state size of 16 and inner dimension of 4096.

\begin{figure}[!t]
    \centering
    \includegraphics[width=1\linewidth]{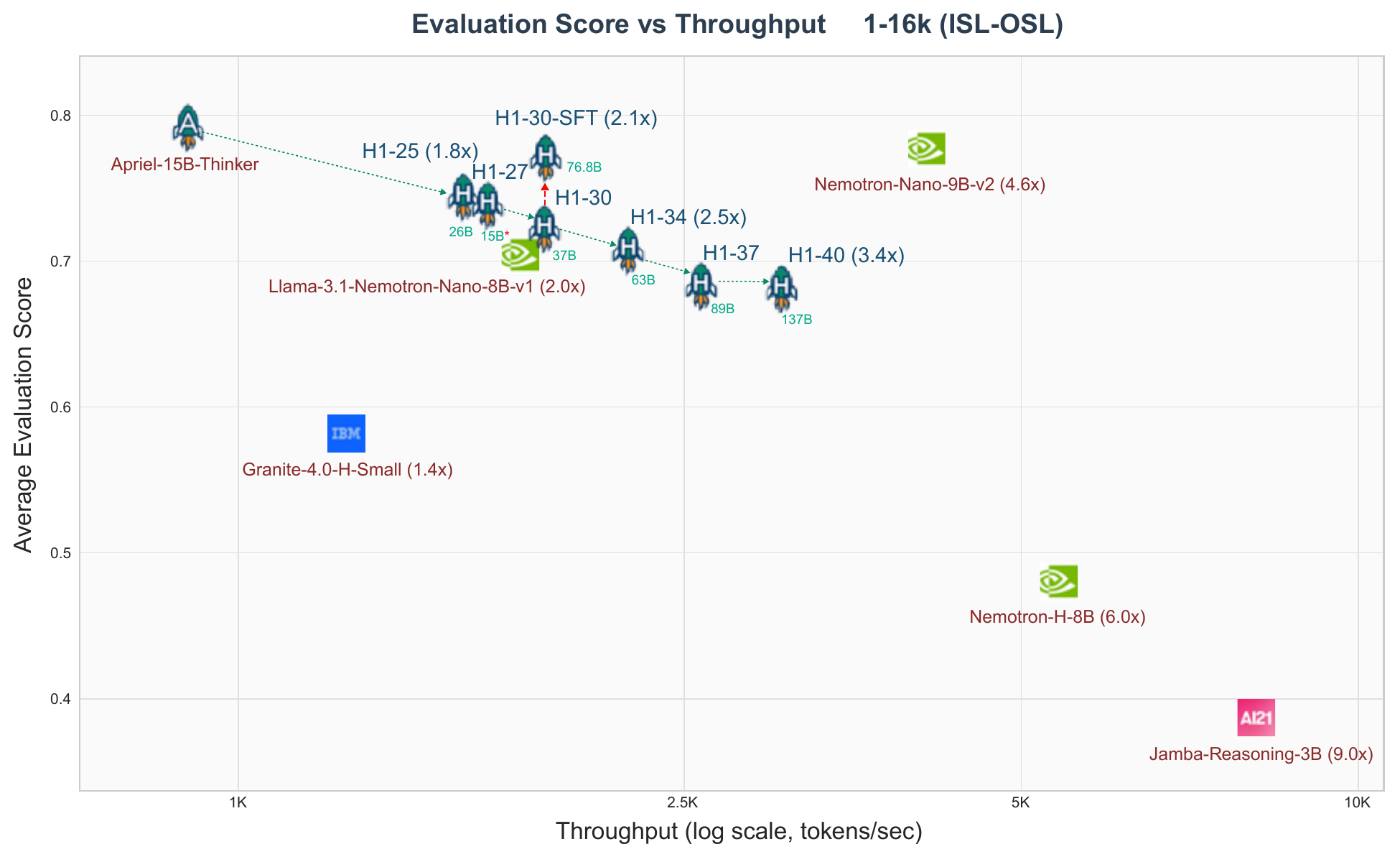}
    \caption{Performance vs. Throughput trade-off for \emph{Apriel-H1-15B-Thinker} variants (H1-25 to H1-40), Apriel-15B-Thinker (transformer) and other open-weights hybrids measured using vLLM backend. We reveal the total number of tokens used to obtain each of the Apriel-H1 models from the Apriel-15B-Thinker below each of the models. $\textcolor{red}{^*}$ H1-27 model has less tokens than H1-25 one because because it used an earlier checkpoint of H1-25 model as its starting point than the checkpoint plotted here for the H1-25.  Apriel-H1-30/50-Thinker-SFT -- H1-30-SFT, is a version of the post-distillation H1-30 model further fine-tuned on a dataset of high-quality reasoning traces and \textit{merged} with a version of H1-30 model (after 55.9B tokens of distillation). The linear decay in model performance as the number of SSM mixers is increased highlights a smooth trade-off between speed and performance. We also include the most recent Nemotron-Nano-9B-v2 model\cite{basant2025nvidia}, that dominates the Pareto frontier in this plot. Importantly, this model has been obtained through direct pre-training of 12B hybrid (20T tokens) on high-quality data, and subsequently post-training using SFT and GRPO phases, and an additional punning phase where the model was pruned from 12B to 9B. We do not include other existing SSM-Transformer hybrid reasoners like M1~\cite{wang2025m1} and Jet-Nemotron~\cite{gu2025jet} due to their small size and absent implementation in the recent vLLM version, we also did not include other models that we could not fit on a single H100 GPU for inference in the 1 input and 16k output tokens scenario.}
    \label{fig:performance}
\end{figure}

\section{Benchmarking}
\label{Benchmarking}

\paragraph{Distilled Apriel-H1 hybrids largely maintain capabilities of the teacher model.}
The goal of our distillation process is to preserve the teacher model’s performance across reasoning-intensive benchmarks such as MATH500 and GSM8k, advanced and olympiad-level mathematics tasks including AIME’24, AIME’25, and AMC23, coding with MBPP, expert-level scientific reasoning and knowledge with GPQA \cite{rein2024gpqa}, as well as instruction following -- IFEval \cite{zhou2023instruction}, and multi-turn dialogue -- MT-Bench \cite{bai2024mt}). As visualized in Fig.~\ref{fig:eval_hybrid_transformer}, we achieve this goal with our best hybrid Apriel-H1-30/50-15B-Thinker-SFT that doubles the throughput with only minimal performance drops.

\paragraph{Comparison to other models in a similar model class.}
In Fig.~\ref{fig:performance} we plot performance and the vLLM output token generation throughput of the Apriel-H1 models, their teacher Apriel-Nemotron-15B-Thinker and other open-weight hybrids that could be run on a single H100 GPU (80GB) (1 input and 16k output tokens). We observe a clear and smooth trade-off between performance and throughput as the number of SSM mixer layers increases within the Apriel-H1 series. Introduction of additional SSM layers reduces raw evaluation scores but leads to significant efficiency improvements—achieving up to 3.4× higher throughput (Apriel-H1-40) compared to the pure Transformer baseline. The post-distillation Apriel-H1 models maintain a favourable position on the performance-throughput plot outperforming other open-weight hybrids such as Llama-3.1-Nemotron-Nano-8B-v1~\cite{bercovich2025llamanemotronefficientreasoningmodels} and Granite-4.0-H1-Small~\cite{granite2025}. We believe that the performance drop of the post-distillation models can be circumvented by further supervised or reinforcement learning fine-tuning on high quality data. This is demonstrated by with the Apriel-H1-30/50--15B-Thinker-SFT (H1-30-ST) variant, which benefits from continued supervised fine-tuning on high-quality reasoning traces nearly matching the teacher’s average score while doubling its throughput.

\paragraph{Critical analysis.}
These results collectively demonstrate that distilling high-capacity reasoning Transformers into hybrid SSM architectures can produce models that preserve core reasoning capabilities while achieving substantially improved inference efficiency. However, the number of training tokens required for this distillation process is notably high—significantly exceeding token budgets typically reported in prior works on base-model distillation~\cite{bick2025llamba,wang2024mamba,goldstein2025radlads}. Although this figure remains small relative to the total compute and data requirements of full-scale pretraining, it nonetheless suggests that reasoning-focused distillation may demand greater data exposure to effectively transfer complex multi-step reasoning behaviours from teacher to student.

Finally, Nemotron-Nano-9B-v2~\cite{basant2025nvidia} defines the current Pareto frontier among open-weight hybrids. This advantage can be attributed to its large-scale, from-scratch pre-training as an SSM-Transformer hybrid on 20 trillion tokens, followed by extensive multi-stage post-training (SFT + GRPO) and a pruning phase that further enhances efficiency. While this demonstrates the upper bound of what direct hybrid pre-training can achieve, it also highlights the substantial compute and data requirements associated with such an approach—resources often beyond the reach of most research or production pipelines. Moreover, pretraining directly as a hybrid remains a risky endeavor, as attention-based architectures constitute a mature and well-understood backbone with predictable optimization dynamics, whereas large-scale hybrid pretraining is still relatively uncharted. Our results therefore suggest that, although large-scale hybrid pretraining is a promising long-term direction, distillation offers a more practical and controllable pathway toward hybrid efficiency without incurring prohibitive resource costs.

\section{Conclusion}
In this technical report, we introduced the Apriel-H1 family of hybrid large language models that combine transformer attention with Mamba sequence mixers through staged, reverse-KL distillation. Our findings demonstrate that high-capacity reasoning transformers can be efficiently converted into hybrid SSM-Transformer architectures while retaining strong reasoning performance. The resulting models achieve up to 3.4× higher inference throughput with minimal degradation on reasoning-heavy benchmarks such as MATH500, GSM8k, AIME, and GPQA.

Our experiments highlight that while distillation provides a stable and controllable pathway toward hybridization, it incurs higher token usage than typical base-model distillation, suggesting that effective transfer of multi-step reasoning may require greater data exposure. In contrast, large-scale from-scratch hybrid pretraining—as exemplified by Nemotron-Nano-9B-v2—can push the Pareto frontier further but demands orders of magnitude more compute and data, and remains a risky and less predictable strategy compared to distillation from an established transformer teacher as for now.

Looking forward, we plan to explore improved linear mixers (e.g., DeltaNet, Gated-DeltaNet) and adaptive hybridization schedules guided by per-layer importance and principled architecture search algorithms to further improve the performance-efficiecy trade-off of distilled linear hybrids .



\printbibliography
\newpage
\section{Contributions and Acknowledgments}
\label{sec:contributors}

\textbf{Core Contributors}

Oleksiy Ostapenko, Luke Kumar, Raymond Li, Denis Kocetkov, Joel Lamy-Poirier, Torsten Scholak

\vspace{1em}

\textbf{Contributors}

Shruthan Radhakrishna, Soham Parikh, Shambhavi Mishra

\vspace{1.25em}

\textbf{Leads \& Management}

\begin{tabularx}{\linewidth}{|>{\raggedright\arraybackslash}p{0.45\linewidth}|>{\raggedright\arraybackslash}X|}
\hline
Srinivas Sunkara & VP, Applied Research \\
\hline
Valérie Bécaert & VP, AI Research \\
\hline
Sebastien Paquet & Research Manager \\
\hline
Sathwik Tejaswi Madhusudhan & Technical co-lead \\
\hline
Torsten Scholak & Technical co-lead \\
\hline
\end{tabularx}








\end{document}